%% file: root.tex
\documentclass[letterpaper, 10 pt, conference]{ieeeconf}  

\IEEEoverridecommandlockouts                              

\usepackage{graphicx}
\usepackage{booktabs}
\usepackage{multirow}
\usepackage{array}

\input{macros}

\title{\LARGE \bf
Real-Time Truly-Coupled Lidar-Inertial Motion Correction and Spatiotemporal Dynamic Object Detection
}

\author{Cedric Le Gentil, Raphael Falque, and Teresa Vidal-Calleja
\thanks{All the authors are with the Robotics Institute at the University of Technology Sydney, Australia.
        {\tt\small \{cedric.legentil;raphael.falque;}\newline \tt\small {teresa.vidalcalleja\}@uts.edu.au}}%
\thanks{Cedric and Teresa are supported by the Australian Research Council Discovery Project under Grant DP210101336.}
\thanks{© 2024 IEEE.  Personal use of this material is permitted.  Permission from IEEE must be obtained for all other uses, in any current or future media, including reprinting/republishing this material for advertising or promotional purposes, creating new collective works, for resale or redistribution to servers or lists, or reuse of any copyrighted component of this work in other works.}
}

\begin{document}

\maketitle
\thispagestyle{empty}
\pagestyle{empty}

\begin{abstract}

Over the past decade, lidars have become a cornerstone of robotics state estimation and perception thanks to their ability to provide accurate geometric information about their surroundings in the form of 3D scans.
Unfortunately, most of nowadays lidars do not take snapshots of the environment but \textit{sweep} the environment over a period of time (typically around 100 ms).
Such a rolling-shutter-like mechanism introduces motion distortion into the collected lidar scan, thus hindering downstream perception applications.
In this paper, we present a novel method for motion distortion correction of lidar data by tightly coupling lidar with Inertial Measurement Unit (IMU) data.
The motivation of this work is a map-free dynamic object detection based on lidar.
The proposed lidar data undistortion method relies on continuous preintegrated of IMU measurements that allow parameterising the sensors' continuous 6-DoF trajectory using solely eleven discrete state variables (biases, initial velocity, and gravity direction).
The undistortion consists of feature-based distance minimisation of point-to-line and point-to-plane residuals in a non-linear least-square formulation.
Given undistorted geometric data over a short temporal window, the proposed pipeline computes the spatiotemporal normal vector of each of the lidar points.
The temporal component of the normals is a proxy for the corresponding point's velocity, therefore allowing for learning-free dynamic object classification without the need for registration in a global reference frame.
We demonstrate the soundness of the proposed method and its different components using public datasets and compare them with state-of-the-art lidar-inertial state estimation and dynamic object detection algorithms.

\end{abstract}

\section{Introduction}

\begin{figure}
    \centering
    \def\imgheight{3.5cm}
    \def\hspace{0.0cm}
    \def\legendspace{0.1cm}
    \def\textsize{\small}
    \begin{tikzpicture}
        \tikzstyle{legend} = [minimum height = 1em, text width = 6.5em,  minimum width = 6.5em, align = center, node distance = 5em, execute at begin node=\setlength{\baselineskip}{8pt}]
        
        \node (a) {\includegraphics[clip,height=\imgheight, trim= 1.3cm 0.8cm 9.5cm 0cm]{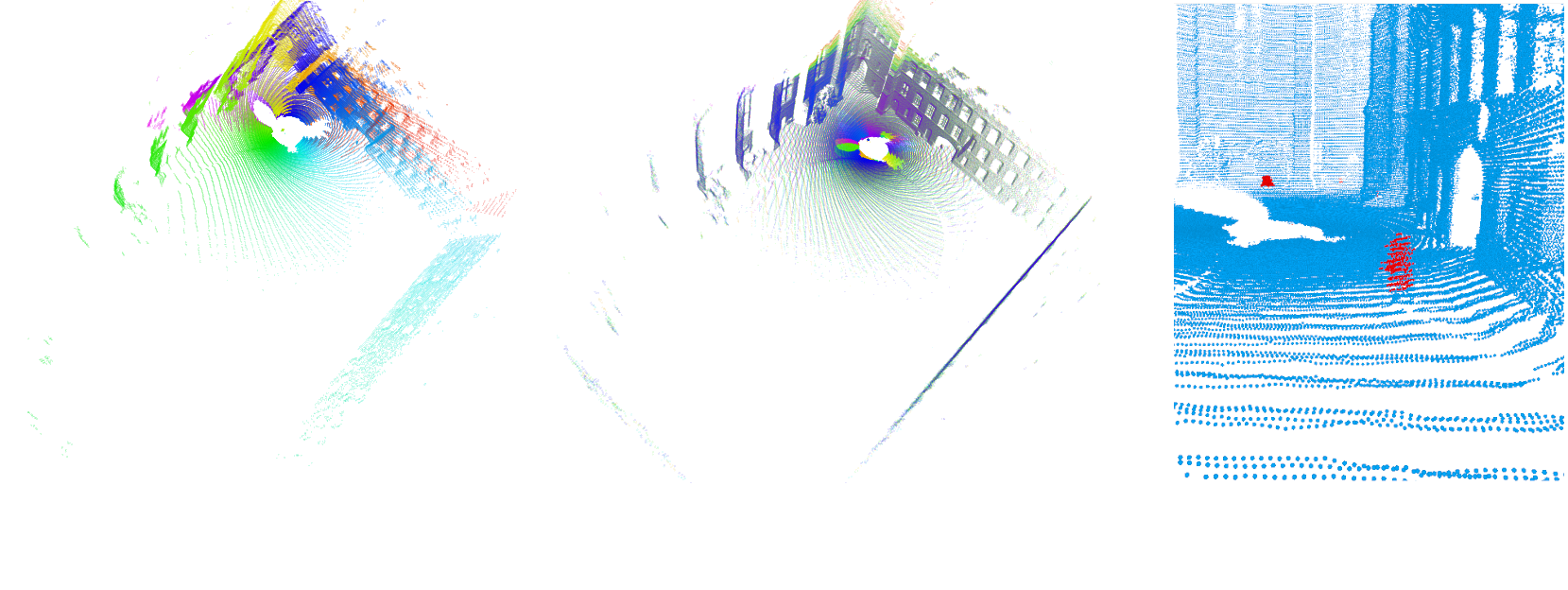}};
        \node[right= \hspace of a] (b) {\includegraphics[clip,height=\imgheight, trim= 6.5cm 1cm 4.5cm 0cm]{figures/tease.png}};
        \node[right= \hspace of b] (c) {\includegraphics[clip,height=\imgheight, trim= 10.5cm 1cm 0.5cm 0cm]{figures/tease.png}};

        \node[legend, below=\legendspace of a]{\textsize (a) Raw lidar\\data};
        \node[legend, below=\legendspace of b]{\textsize (b) Motion-corrected data};
        \node[legend, below=\legendspace of c]{\textsize (c) Dynamic object detection};
    \end{tikzpicture}
    \caption{The proposed method introduces a real-time IMU-lidar motion distortion algorithm that undistorts lidar data (b) to later perform dynamic object detection (c). The colour in (a) and (b) depicts the points' timestamps, while to corresponds the classification output in (c).}
    \label{fig:teaser}
\end{figure}

A key condition for popularising autonomous systems in society is their ability to robustly perceive their environment and safely interact with humans and other moving agents in their surroundings.
Recent advancements in computer vision have provided the robotics community with unprecedented perception abilities.
Many of these methods are based on visual images and deep neural networks.
However, standard cameras have limitations (e.g., low illumination scenes), not all robotic systems can include a camera (e.g., privacy), and the computational burden of some deep neural networks can be prohibitive for embedded systems.
Additionally, these methods sometimes fail to generalise to unseen scenes \cite{sunderhauf2018limits}, and for some critical industries, they lack explainability \cite{willers2020safety}.
As illustrated in Fig.~\ref{fig:teaser}, this paper focuses on two components of lidar-inertial perception: an ego-motion estimation approach and a dynamic object detection algorithm.
This work is based on traditional geometric models and does not require any training phase.

Due to the sweeping nature of most of nowadays lidars, the scans collected from a moving sensor suite are subject to motion distortion.
Throughout the years, various techniques have been proposed to address this distortion and perform state estimation.
A common approach is to combine lidar and inertial data from an \ac{imu} into a single algorithm \cite{lee2024lidar}.
The idea is that the inertial information collected during a lidar scan can be used to correct the motion that happened during the scan.
Most of the popular frameworks available for lidar-inertial state estimation rely on an ``open-loop" motion distortion correction of the lidar scans in the sense that it is not part of their estimation algorithm, but a preprocessing step that often relies on the current knowledge of the sensor's pose and velocity~\cite{Ye2019}.
Accordingly, such a method needs roughly known initial conditions to perform optimally.
Thus, the claim for ``tightly-coupled" lidar-inertial state estimation is only partially true~\cite{shan2020liosam}.
Our method proposes a truly coupled technique that does not require any initialisation or specific initial conditions.
This initialisation-free undistortion is motivated by the task of dynamic object detection without the need for global registration of the lidar data, opening future opportunities for robust lidar-inertial odometry and mapping framework in dynamic environments.

The main idea of the proposed motion distortion correction method is to use our previous work on continuous preintegrated measurements from \cite{legentil2023latent} between timestamps $\frametime{0}$ and $\frametime{1}$ as the 6-\ac{dof} continuous trajectory representation over the same temporal window.
Doing so, based on the assumption of constant bias between $\frametime{0}$ and $\frametime{1}$, the trajectory is fully defined by the value of the \ac{imu} biases (6 values), the velocity at time $\frametime{0}$ (3 values), and the orientation of the gravity vector at time $\frametime{0}$ (2 values).

The ability to robustly undistort lidar data is a key condition for model-based scene understanding.
In this work, we integrate the proposed motion correction method with our previous work on dynamic object detection \cite{falque2023dynamic}.
In this previous work, we relied on known poses of the sensor in a global reference frame to extract the spatiotemporal normal vector of the points in the scene.
The known sensor poses were provided by the known kinematics of a robotic arm or the output of a visual-inertial framework.
In this paper, we show that it is possible to reliably classify the dynamic objects in the scene using only lidar and inertial data\footnote{Most of the lidars available today include an \ac{imu}.} by performing motion correction over a short period of time and analysing the spatiotemporal normal vectors without the need for global registration of the lidar data.
It makes our method robust to any estimation drift, which contrasts with methods like \cite{schmid2023dynablox} that require known poses and maintain a global map of the environment.

To summarise, the contributions of this work are as follows: the derivation of a real-time algorithm for initialisation-free motion distortion correction based on lidar-inertial data; its integration with an explainable model-based dynamic object detection; the evaluation of the proposed framework components with real world-data; and the open-source release of the code and dataset with ground truth annotation used for validation\footnote{\href{https://uts-ri.github.io/lidar_inertial_motion_correction/}{https://uts-ri.github.io/lidar\_inertial\_motion\_correction/}}.

\section{Related work}


\subsection{Lidar inertial state estimation}

Provided the ubiquity of \acp{imu} and the popularisation of 3D lidars, a lot of work has focused on the problem of lidar-inertial odometry and state estimation.
More than a decade ago, the authors of \textit{Zebedee}~\cite{Bosse2012} introduced a lidar-inertial 3D mapping device that relied on a piece-wise linear motion model as a continuous representation of the sensor trajectory to address the challenging motion of the 2D lidar used.
Later, \textit{LOAM} \cite{Zhang2015} was presented.
It introduced a certain amount of concepts that are still the cornerstone of many lidar-inertial frameworks proposed nowadays (including ours).
Its formulation relies on scan-to-scan and scan-to-map lidar registration using lidar features corresponding to planar patches and edges in the environment.
Despite its ability to use inertial data, \textit{LOAM} algorithm uses lidar and \ac{imu} data in a fairly loose manner in the sense that it was mostly used as an open-loop undistortion step and an educated initial guess for the scan-to-scan matching.

Some works attempted to couple the \ac{imu} and lidar data by using a factor graph formulation \cite{Ye2019,shan2020liosam} or in an EKF \cite{xu2022fastlio2}.
Recently, \textit{DLIO} \cite{chen2023dlio} introduced a continuous-time undistortion mechanism based on a constant-jerk motion model to interpolate the inertial predictions between \ac{imu} measurements.
Despite the claim of ``tight" coupling, these methods use the \ac{imu} to correct the motion distortion in a one-off state-propagation pre-processing step based on the last state estimate.
It means that the position of the lidar during a lidar scan is not estimated using both geometric and inertial information.
Therefore, the undistortion accuracy is tied to the accuracy of the previous scan-matching steps.
On another hand, methods with a continuous state representation \cite{Bosse2012,Tang2019} perform ``non-rigid" registration of the lidar scans by considering the points independently.
A downside of this approach is the size of the estimated state that inherently yields a heavier computation burden in general, as shown in \cite{cioffi2022continuous}.

The concept of preintegration, originally presented in \cite{Lupton2012}, corresponds to the creation of pseudo measurements that combine the information from multiple \ac{imu} measurements without the need to know the initial conditions.
In the context of lidar-\ac{imu} calibration, the work in \cite{LeGentil2018} relied on a continuous representation of the inertial data to virtually upsample the \ac{imu} measurement and perform preintegration, thus providing inertial information for each lidar point.
This has been later extended with \textit{IN2LAAMA} \cite{LeGentil2021}, an offline localisation and mapping framework.
With novel continuous preintegrated measurements \cite{LeGentil2021b} then \cite{legentil2023latent}, \textit{IN2LAAMA} is creating a bridge between discrete and continuous-time state representations with discrete pose, velocity and bias variables for each lidar scan but locally considering continuous trajectories within each of the scans, thus allowing for ``non-rigid" scan-to-scan registration.
In this paper, we push this notion further by eliminating the concept of scan and removing the pose from the estimated state.

\subsection{Dynamic object detection}

As mentioned earlier, a common approach to detect dynamic objects is to leverage semantic segmentation of geometric sensor data.
The classic strategy is to use a UNET architecture to classify all the pixels/3D points related to potential dynamic elements.
In \cite{chen2019suma++}, the authors proposed to use such a method with the \ac{cnn} Rangenet++~\cite{milioto2019rangenet++} and the KITTI Vision Benchmark dataset~\cite{geiger2012we}.
This work was later extended by changing the network input space for a stream of consecutive scans~\cite{chen2021moving}.
Alternatively, if depth and \ac{rgb} images are acquired simultaneously, the \ac{rgb} data can be used with a standard semantic segmentation network, such as a \ac{rcnn}, and then mapped back into the depth information for classifying potential dynamic objects~\cite{henein2020dynamic}.

As mentioned above, these algorithms can classify \emph{potential} dynamic objects.
However, they can not separate moving elements from static ones that have the potential to be dynamic (e.g., a parked and a moving vehicle).
For this reason, some of the literature focuses on detecting dynamic objects regardless of the semantic information.
A straightforward approach is to integrate dynamic object detection into an occupancy grid mapping framework.
The occupancy grid can then be used to detect areas that used to be \emph{free} and are now \emph{occupied} (i.e., all the newly occupied areas are parts of dynamic objects).
This strategy has been proposed in Dynablox~\cite{schmid2023dynablox}, where the authors build upon the voxel occupancy map from the Voxblox framework~\cite{oleynikova2017voxblox} to solve the dynamic object detection.
Similarly, Mersch et al. proposed a framework that segments dynamic objects by looking at the discrepancies between a lidar scan and the map. They propose to use a sparse 4D (i.e., [x, y, z, time]) \ac{cnn} to segment points that are dynamic~\cite{mersch2023ral}.

In recent work, we introduced a simple method to detect dynamic elements within range scan data as a byproduct of the spatiotemporal normals computation~\cite{falque2023dynamic}.
While not requiring to solve the mapping problem as in ~\cite{schmid2023dynablox, mersch2023ral}, our previous work relied on the accurate knowledge of the sensor poses in a global reference frame.
By, integrating~\cite{falque2023dynamic} over the short temporal windows of the proposed lidar-inertial undistortion method, we demonstrate the ability to recover object dynamicity based on spatiotemporal normal computation without the need for global registration.

\section{Method}

\subsection{Preintegration background}

The preintegration concept allows for the aggregation of \ac{imu} measurements into pseudo measurements that relate the inertial pose at different timestamps.
In the context of motion distortion correction we are only interested in the relative motion between timestamps $\frametime{0}$ and $\frametime{1}$ considering the reference frame of the \ac{imu} at time $\frametime{0}$ as the ``local" reference frame. We denote it as $\refframe{\frametime{0}}{I}$.
Accordingly, the pose of the \ac{imu} at $\frametime{0}$ is the identity.
In such conditions, the standard preintegration equations that involve the preintegrated measurements $\drot{\frametime{0}}{\time{}}$ (rotation) and $\dpos{\frametime{0}}{\time{}}$ (position) can be simplified as
\begin{align}
    \rot{I_0}{\time{}} &= \drot{\frametime{0}}{\time{}}
    \label{eq:preint_rot}
    \\
    \pos{I_0}{\time{}} &= (\time{} - \frametime{0})\vel{I_0}{\frametime{0}} + \frac{(\time{} - \frametime{0})^2}{2}\gravity_{I_0} + \dpos{\frametime{0}}{\time{}},
    \label{eq:preint_pos}
\end{align}
where $\gravity_{I_0}$ is the gravity vector in $\refframe{\frametime{0}}{I}$.
The notations $\rot{I_0}{t}$, $\vel{I_0}{t}$, and $\pos{I_0}{t}$ represent the \ac{imu} orientation, velocity, and position in $\refframe{\frametime{0}}{I}$, respectively.

For the rest of the paper, we will use the homogenous transformation $\T{I_0}{t}$ to refer to the \ac{imu} pose at time $t$
\begin{equation}
    \T{I_0}{t} = \left[\begin{smallmatrix}
        \rot{I_0}{t} & \pos{I_0}{t} \\ \mathbf{0} & 1
    \end{smallmatrix}\right].
    \label{eq:preint_pose}
\end{equation}
Note that the preintegrated measurements in \eqref{eq:preint_rot} and \eqref{eq:preint_pos} are considered to be bias-free.
However, in real-world scenarios, the inertial measurements are affected with slow-varying additive biases.
Following the seminal work in~\cite{Lupton2012}, most preintegration methods offer a Taylor expansion of the preintegrated measurements to enable post-integration bias correction.
Thus the \ac{imu} pose \eqref{eq:preint_pose} is not only a function of the preintegrated measurement, the initial velocity, and gravity vector, but it is also a function of the accelerometer and gyroscope biases.
For more details about \ac{imu} preintegration and bias correction mechanisms please refer to~\cite{legentil2023latent}.

\subsection{Motion correction}

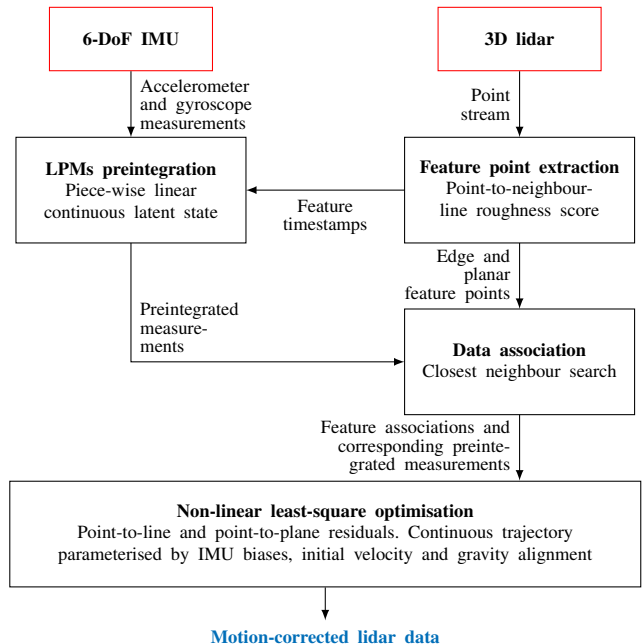
\begin{figure}
    \centering
    \input{figures/overview_undistortion}
    \caption{Overview of the proposed lidar-inertial undistortion pipeline. The sensors' trajectory is modelled with continuous preintegrated measurements \cite{legentil2023latent} resulting in an estimated state of only 11-DoF.}
    \label{fig:overview_diagram}
\end{figure}
To detect moving objects in the lidar data, we first need to correct for the motion distortion due to the ego-motion of the sensor.
Our method relies on both a lidar and a 6-\ac{dof} \ac{imu} to provide distortion-free point clouds of the sensor's surroundings based on a non-linear least-square minimisation of distances between lidar points.
An overview of the proposed approach is present in Fig.~\ref{fig:overview_diagram}.
We consider data collected between timestamps $\frametime{0}$ and $\frametime{1}$ by lidar and \ac{imu} rigidly mounted with $\Tc$ their extrinsic calibration.
Please note that the proposed method does not rely on the traditional notion of ``lidar scan" as the points are considered independently.
However, the time window [$\frametime{0}$, $\frametime{1}$] needs to be long enough to allow for data association as mentioned later in this section.

Given the raw data with timestamps, feature points are extracted from the lidar data.
These correspond to planar and edge points.
Then, preintegrated measurements are computed for each lidar feature following the \acp{lpm} introduced in \cite{legentil2023latent}.
Using \eqref{eq:preint_pose}, it is possible to project any lidar point into the sensor's referential frame at time $\frametime{0}$ where data association between features can be performed.
Each data association lead to a point-to-plane or point-to-line residual (depending on the nature of the features) in our non-linear least-square optimisation.
Using the Levenberg-Marquardt algorithm we estimate the state $\state = \begin{bmatrix} \biasacc{}& \biasgyr{} & \vel{I_0}{\frametime{0}} & \gravity_{I_0} \end{bmatrix}$, with $\biasacc{}$ and $\biasgyr{}$ the constant accelerometer and gyroscope biases, $\vel{\frametime{0}}{}$ the \ac{imu} velocity at time $\frametime{0}$, $\gravity_{I}$ the gravity vector in the \ac{imu} reference frame at time $\frametime{0}$.

\subsubsection{Feature detection}

\begin{figure}
    \centering
    \includegraphics[width=0.8\columnwidth]{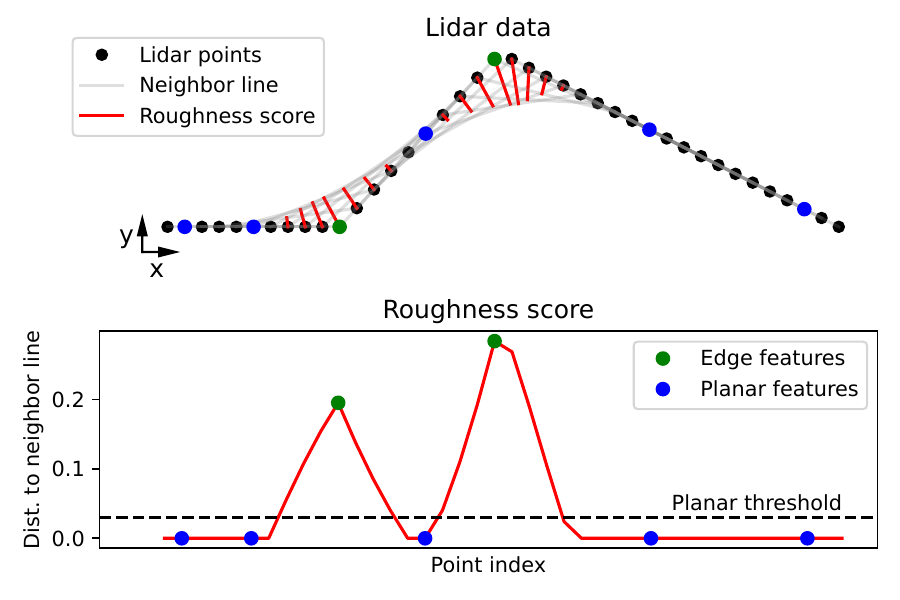}
    \vspace{-0.5cm}
    \label{fig:feature_detection}
    \caption{Illustration of the proposed lidar feature detection as the point-to-line distance with the ``neighbour line".}
\end{figure}

Let us denote $\point{L}{i}$ a lidar point collected at time $\time{i}$ with $\frametime{0} \leq \time{i} \leq \frametime{1}$.
Most of the lidars available nowadays collect data by swiping the environment with a fixed number of laser beams.
Let us use the term ``channel" to refer to the successive points collected by one of the lasers.\footnote{For example, there are 16 channels in a Velodyne VLP16, while a Blickfeld Cube1 or a Livox Avia only have 1 channel.}
The proposed feature detection computes a roughness score for each point in a channel as\footnote{For the sake of notation simplicity, the point indices in \eqref{eq:roughness_score} reflect the case with only 1 channel.}
\begin{equation}
    r(\point{L}{i}) = \textstyle \frac{\left\Vert(\point{L}{i}-\point{L}{i-n})\times(\point{L}{i}-\point{L}{i+n})\right\Vert}{\left\Vert\point{L}{i-n}-\point{L}{i+n}\right\Vert},
    \label{eq:roughness_score}
\end{equation}
with $n$ a user-defined parameter.
This score corresponds to the point-to-line distance between the point under consideration and the line defined by neighbouring points from the same channel.
Points with a score under a certain threshold are considered as planar features.
To limit the computational burden in the rest of the proposed method, only a random subset of these planar features is used if too many points are planar.
The edge features correspond to local maxima in the roughness score that is above the planar threshold as illustrated in Fig.~\ref{fig:feature_detection}.
Let us use the generic notation $\feature{L}{i}$ to refer to selected feature points regardless of their nature.

\subsubsection{Data association}

As later illustrated in Fig.~\ref{fig:temporal_graph}, we divide the time window into $K$ ``segments" of the same duration.
Each segment requires a spatial overlap with the other segments.
Accordingly, the time difference between $\frametime{1}$ and $\frametime{0}$ must be superior to $K$ times the duration of the scanning pattern of the lidar used (e.g., with $K=2$, $\frametime{1} - \frametime{0}>200\units{ms}$ for a standard spinning lidar operating at $10\units{Hz}$).
The data association between two successive segments consists first of the following steps:
\begin{itemize}
    \item Projection of all the features to the \ac{imu} frame at time $\frametime{0}$ leveraging \eqref{eq:preint_pose}:
    $
    \left[\begin{smallmatrix}
        \feature{I_0}{i}\\
        1
    \end{smallmatrix}\right]
    = \T{I_0}{\time{i}}(\state) \Tc \left[\begin{smallmatrix}\feature{L}{i}\\ 1\end{smallmatrix}\right]
    $.
    \item Building KDTrees \cite{Bentley1975} with the features of each segment (independently for each feature type).
    \item Closest neighbour search between segment pairs (2 neighbours for edges, and 3 for planar features).
\end{itemize}
We consider an association of a feature with its neighbours as valid if the Euclidean distance separating them is under a user-defined threshold.
We denote $\asso{i} = \{\feature{L}{i}, \feature{L}{j}, \feature{L}{k} (,\feature{L}{l}) \}$ one of such associations with $\feature{L}{i}$ the feature under consideration and $\feature{L}{j}$, $\feature{L}{k}$, $($, and $\feature{L}{l})$ the target features.

\subsubsection{Cost function}

Provided the aforementioned data associations, our estimation problem corresponds to the non-linear least-square optimisation of the cost function as 
\begin{equation}
    \state^* = \underset{\state}{\operatorname{argmin}} \  \sum_{\asso{i} \in \assoset} (d(\asso{i}, \state))^2,
    \label{eq:optimisation}
\end{equation}
where $d(\asso{i}, \state)$ is the distance function specific to the type of features.
For edges, the distance function is
\begin{equation}
    d(\asso{i}, \state) = \textstyle \frac{\left\Vert(\feature{I_0}{i}-\feature{I_0}{j})\times(\feature{I_0}{i}-\feature{I_0}{k})\right\Vert}{\left\Vert\feature{I_0}{j}-\feature{I_0}{k}\right\Vert},
\end{equation}
and for planar points
\begin{equation}
    d(\asso{i}, \state) = \textstyle\frac{(\feature{I_0}{i}-\feature{I_0}{j})^\top\left((\feature{I_0}{j}-\feature{I_0}{k})\times(\feature{I_0}{j}-\feature{I_0}{l})\right)}{\left\Vert(\feature{I_0}{j}-\feature{I_0}{k})\times(\feature{I_0}{j}-\feature{I_0}{l})\right\Vert}.
\end{equation}
Note that the state appears on the right-hand side through the transformation of the lidar points into the first \ac{imu} frame.

\subsection{Dynamic object detection}

This subsection provides the technical details of the integration of our previous work~\cite{falque2023dynamic} with the proposed undistortion method.

\subsubsection{Overview}

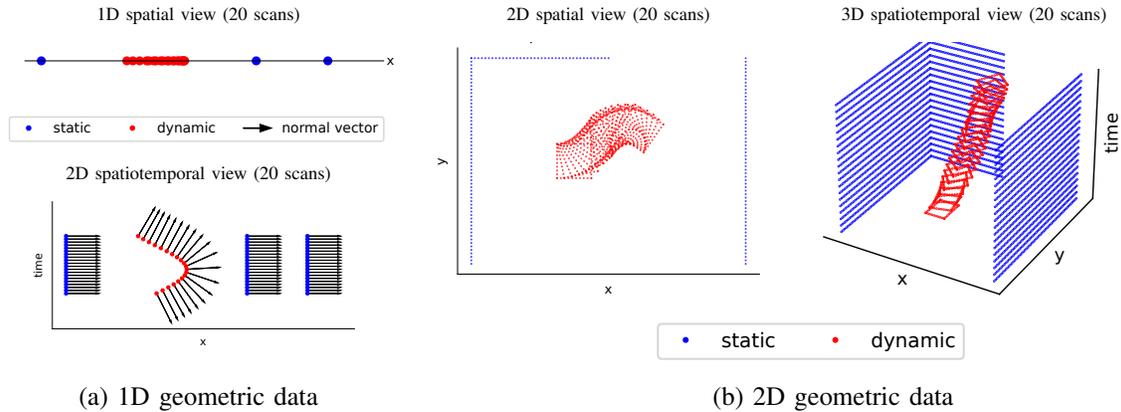
\begin{figure*}
    \centering
    \input{figures/spatiotemporal_view}
    \caption{Intuition for the spatiotemporal normal computation as a proxy for dynamic point detection, 1D in (a) and 2D in (b). The spatiotemporal normal vectors are related to the velocity of an object; for static objects, the temporal component of the normals is null (the normals have been omitted in (b) for the sake of readability)~\cite{falque2023dynamic}.}
    \label{fig:dynamic_intuition}
\end{figure*}


Let us define a dynamic object as an element in the environment that has a non-null velocity with respect to an earth-fixed reference frame.
Thus, the definition of dynamic object here is agnostic to the type or nature of the objects present in the environment.
The goal is to classify the the individual points in the distortion-free lidar data based on a per-point \emph{dynamicity score}.
As illustrated in Fig.~\ref{fig:dynamic_intuition}, the normal vector to the spatiotemporal surface contains information about the velocity of an object.
The temporal component of these spatiotemporal normals is used as our score.
In the 1D scenario, it directly links to the object velocity while it corresponds to the projection of the velocity on the spatial normal vector in higher dimensions.

Concretely, the dynamicity score computation for each point consists of the following steps:
\begin{itemize}
    \item Spatial neighbour search.
    \item Computation of the local spatiotemporal covariance matrix.
    \item Normal computation (eigendecomposition, eigenvector with the smallest eigenvalue).
    \item Score as the temporal component of the normal, and thresholding.
\end{itemize}

\subsubsection{Dynamicity score via spatiotemporal normal}

Let us consider the undistorted lidar points between $\frametime{0}$ and $\frametime{1}$ as $\pointcloud{}{}$.
To provide a score to a point $\point{I_0}{i}$, we first perform a spatial neighbourhood search in $\pointcloud{}{}$.
Defining $\nn_i$ as the set of neighbour points to $\point{I_0}{i}$, the local covariance $\cov_i$ is computed as
\begin{align}
    &\cov_i = {\textstyle\frac{1}{\Vert\nn_i\Vert}}\sum_{\point{I_0}{v} \in \nn_i} \left(\left[\begin{smallmatrix}\point{I_0}{v}\\ \time{v}\end{smallmatrix}\right] - \mean_i\right)\left(\left[\begin{smallmatrix}\point{I_0}{v}\\ \time{v}\end{smallmatrix}\right] - \mean_i\right)^\top
    \\
    &\text{with}\quad \mean_i = {\textstyle\frac{1}{\Vert\nn_i\Vert}}\sum_{\point{I_0}{v} \in \nn_i} \left[\begin{smallmatrix}\point{I_0}{v}\\ \time{v}\end{smallmatrix}\right].
    \nonumber
\end{align}

We define the dynamicity score $\dscore_i^j$ as the absolute value of the temporal component (last component) of the Eigenvector associated with the smallest Eigenvalue of $\cov_i$.
This value is related to the projection of the object velocity in the 3D space along the spatial normal vector at that location.
The point classification as static or dynamic simply consists of thresholding $\dscore_i^j$.
Please note that this concept is not specific to lidar data and can therefore be applied to data from depth-cameras too as shown in the experiment section of this paper as previously shown in \cite{falque2023dynamic}.

\section{Experiment}

\subsection{Implementation}

\begin{figure}
    \centering
    \includegraphics[clip, width=\columnwidth]{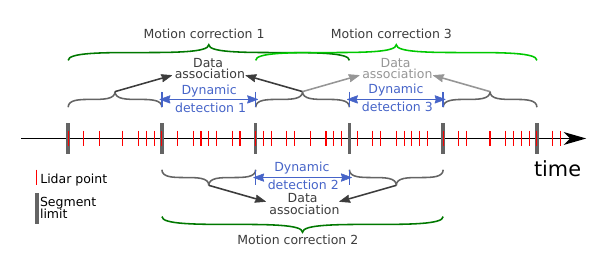}
    \vspace{-0.5cm}
    \caption{Illustration of the proposed pipeline's timeline in terms of lidar data used for motion correction and dynamic object detection. As per our implementation, 3 segments (of $150\units{ms}$) per temporal window are used with a sliding window increment of 1 segment. For the dynamic detection $n$, the 3 segments of the motion correction $n$ are used. In each motion correction, the lidar features from the first and last segments in the temporal window are used for data association.}
    \label{fig:temporal_graph}
\end{figure}

Our open-source C++ ROS (1 and 2) implementation of the undistortion has been made available.\footnote{\href{https://uts-ri.github.io/lidar_inertial_motion_correction/}{https://uts-ri.github.io/lidar\_inertial\_motion\_correction/}. Please note that the released code also integrates an implementation of \cite{legentil2024reverting} for scan-to-map-style odometry and mapping in dynamic environments.}
We rely on \textit{Ceres} \cite{Agarwal} for the non-linear least-square problem in \eqref{eq:optimisation}.
The length of the time window for the motion distortion correction is $450\units{ms}$ with segments of $150\units{ms}$.
The undistortion of all points is performed in a sliding window manner with increments of $150\units{ms}$, as shown in Fig.~\ref{fig:temporal_graph}.
The estimated velocity and gravity orientation of a given window are used as initial guess for the following window.
To lighten the computational load, the \acp{lpm} are computed at a fixed frequency of $1\units{kHz}$ and interpolation is performed for the feature timestamps in between.

The dynamic object detection is implemented by integrating the framework for spatiotemporal normals computation~\cite{falque2023dynamic} consisting of four steps that are, point cloud accumulation, downsampling, dynamic score computation, and upsampling.
Note that here, there is no need for point cloud accumulation as the undistortion step occurs over several scans, thus alleviating the need for global pose.
The parameters have been tuned to work with 
the density of the pointcloud from the dataset.
The downsampling and upsampling modules are based on voxels of size $0.1\units{m}$ and $0.2\units{m}$, respectively.
The threshold for classification of dynamic objects is set to $0.4$ and the search radius for the spatiotemporal normals is set to $0.3m$.
Following \cite{falque2023dynamic}'s notations, $N=1$, $d_r=0.3$, $thr=0.4$.

\subsection{Motion correction}

It is important to note that the proposed motion correction method does not aim at performing global odometry but instead targets local undistortion of the lidar data.
Accordingly, standard benchmarking methods for odometry frameworks, like pose trajectory accuracy, are not considered in this section.
Instead, this evaluation focuses on the consistency of the lidar data after correction against the 3D map of the environment.
We use the \textit{Newer College Dataset} \cite{zhang2021multicamera} as it includes both inertial and geometric data from an \textit{Ouster OS0} (128 beams) lidar and its embedded \ac{imu}, as well as the ground-truth map of the area collected with a survey-grade \textit{Leica BLK360} lidar.
To benchmark our motion correction method we choose to compare with state-of-the-art real-time lidar-inertial estimation frameworks that are \textit{Fast-LIO2} \cite{xu2022fastlio2} and \textit{DLIO} \cite{chen2023dlio} that assert tight coupling of lidar and inertial data in their state estimation pipelines.

\subsubsection{Accuracy}
\begin{figure}
    \centering
    \includegraphics[width=1\columnwidth]{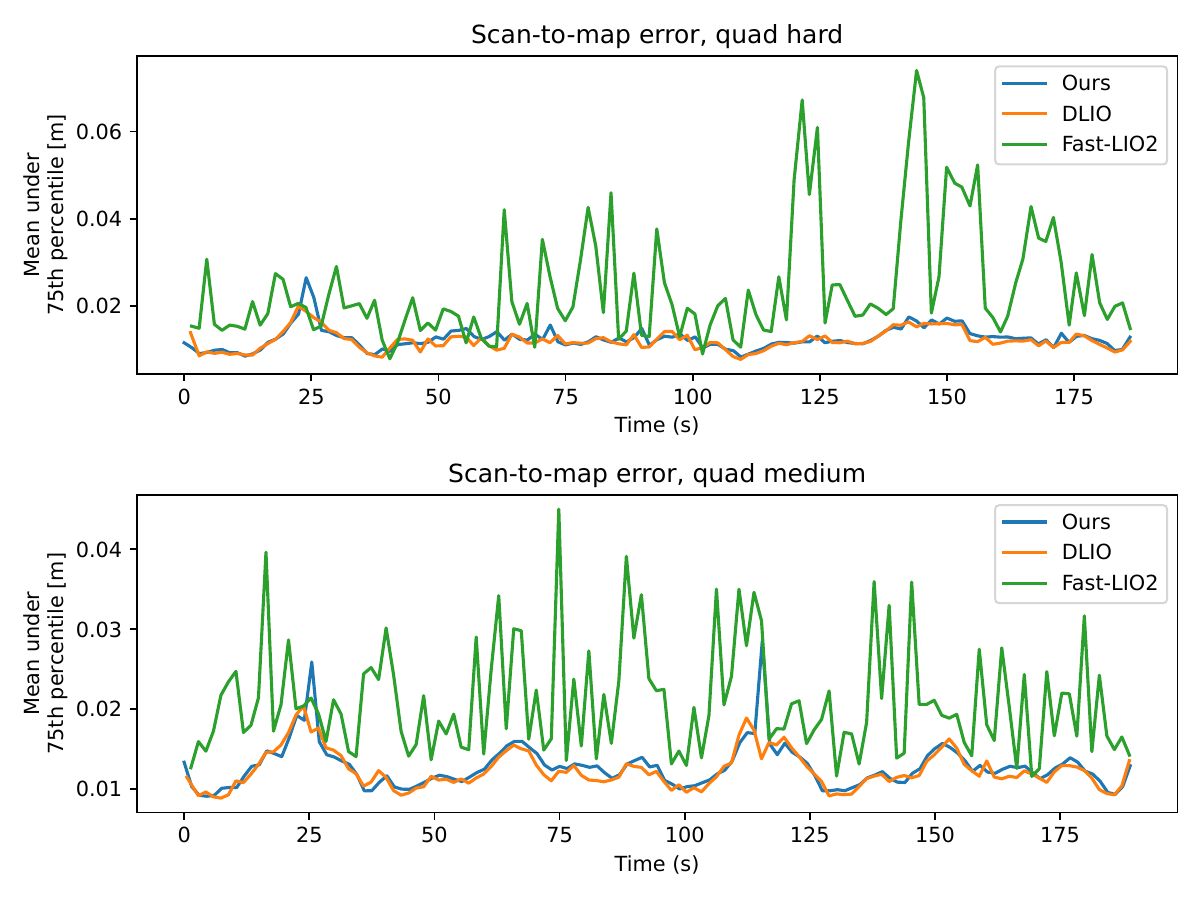}
    \vspace{-0.5cm}
    \caption{Comparison of the motion distortion correction accuracy between the proposed method and state-of-the-art lidar-inertial odometry frameworks. Distance between ground-truth map and motion-corrected lidar data after ICP registration.}
    \label{fig:motion_distortion_results}
\end{figure}

This setup consists of saving to disk all the motion-corrected scans and their associated timestamps before registering them into the ground-truth map using the ground-truth poses and a final ICP alignment.
Once the scans are aligned, we query the smallest distance between each lidar point in the scans and the map.
As the scans and the map do not have a 100\% overlap (some areas seen by the mobile lidar are not observed from any of the survey lidar poses), we compare the mean distances of the 75\% best matches.
Fig.~\ref{fig:motion_distortion_results} shows this mean error through time for the three methods on both the \textit{quad medium} and \textit{quad hard} sequences.

Note that these sequences allow for a fair comparison with both \textit{Fast-LIO2} and \textit{DLIO} because the data start with very little motion allowing for close-to-optimal initialisation of the state.
One can see that \textit{Fast-LIO2} under-performs compared to our method and \textit{DLIO}.
The fact that \textit{DLIO} and the proposed method have similar performance\footnote{The occasional spikes in \textit{Ours} are likely due to a sub-optimal result of the ICP refinement.} with a mean error close to both the value of the sensor's noise ($\approx 0.02\units{m}$) and the map resolution ($0.01\units{m}$ voxel downsampled) confirms that both methods can recover the motion distortion of the lidar accurately.

\subsubsection{Robustness to initial conditions}
In the previous experiments, we have provided the optimal initialisation conditions for \textit{Fast-LIO2} and \textit{DLIO}.
Here we explore the scenario where the sensor suite is moving when switched on.
In other words, in terms of state estimation, it means that the initial conditions of the lidar dynamics are not accurately known.
We compare both our approach and \textit{DLIO} in such scenarios by starting the \textit{Quad hard} sequence at timestamps $120\units{s}$ and $141\units{s}$ (these correspond to higher rotational motion).

\begin{figure}
    \centering
    \def\legendsize{\small}
    \def\imgwidth{0.47}
    \def\vertspace{1cm}
    \def\horispace{0.1cm}
    \def\legendspace{-0.05cm}
    \begin{tikzpicture}
        \node (overall) {\includegraphics[clip, trim=5.5cm 4.5cm 5.5cm 1cm, width=\imgwidth\columnwidth]{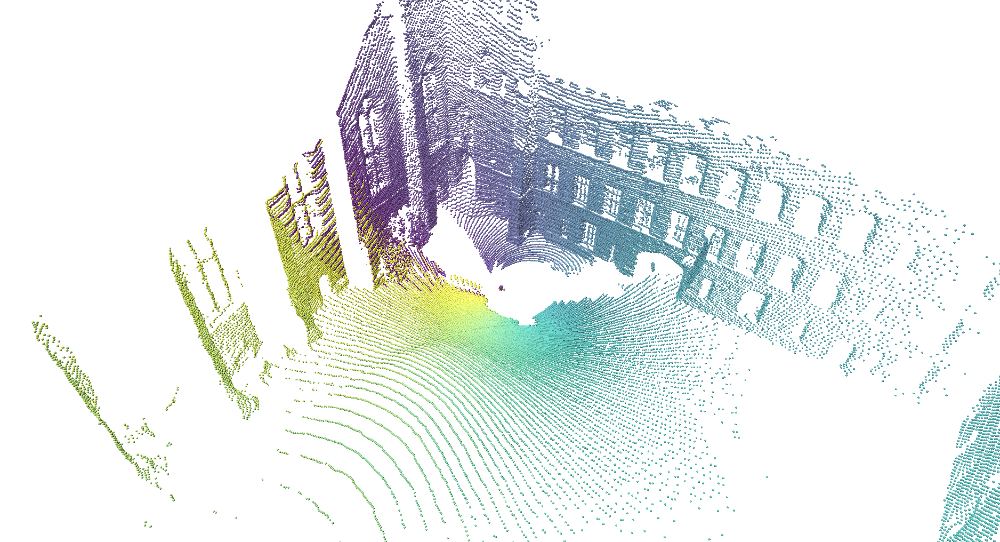}};
        \node[below = \vertspace of overall] (overlap) {\includegraphics[clip, trim=5.5cm 4.5cm 5.5cm 1cm, width=\imgwidth\columnwidth]{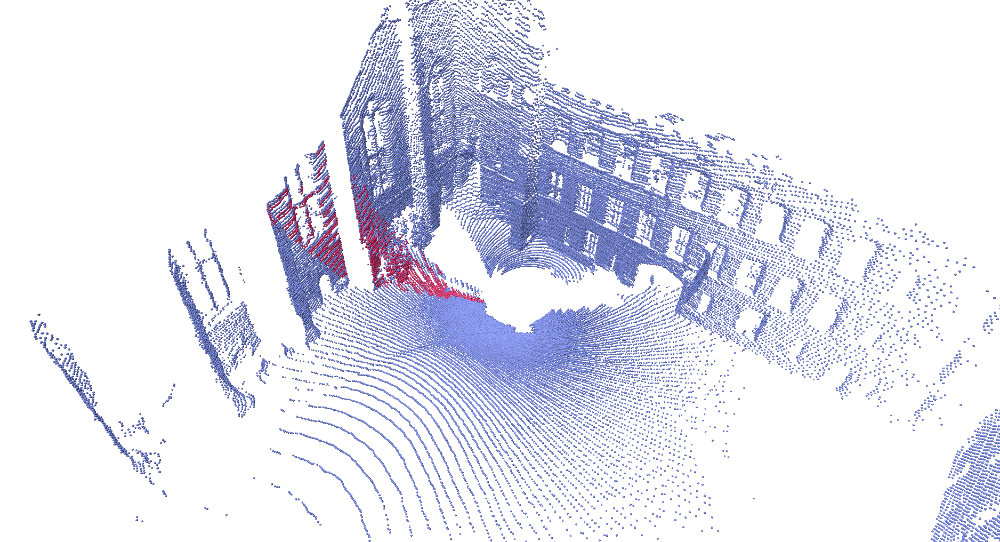}};
        \node[right=\horispace of overall] (ours) {\includegraphics[clip, trim=2cm 0cm 0cm 0cm, width=\imgwidth\columnwidth]{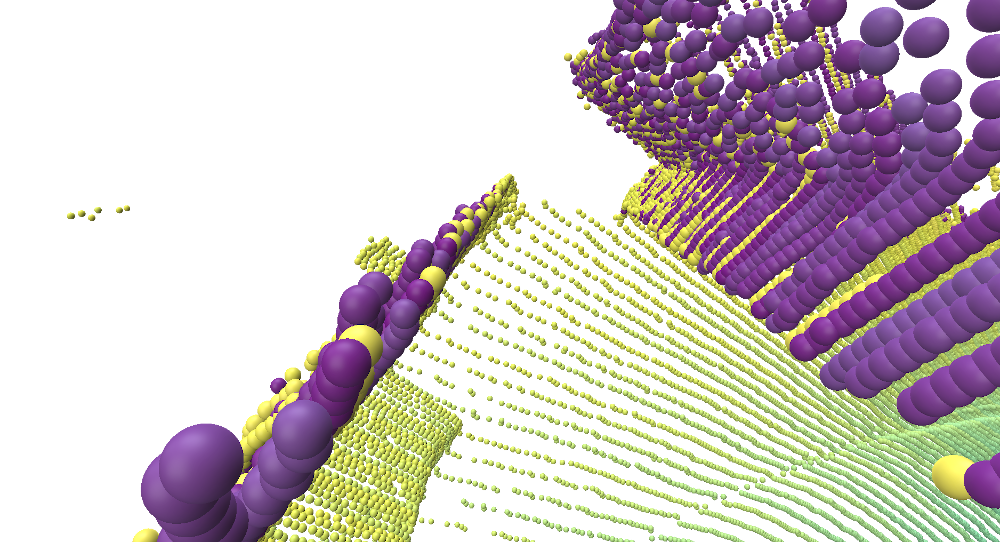}};
        \node[right = \horispace of overlap] (dlio) {\includegraphics[clip, trim=2cm 0cm 0cm 0cm, width=\imgwidth\columnwidth]{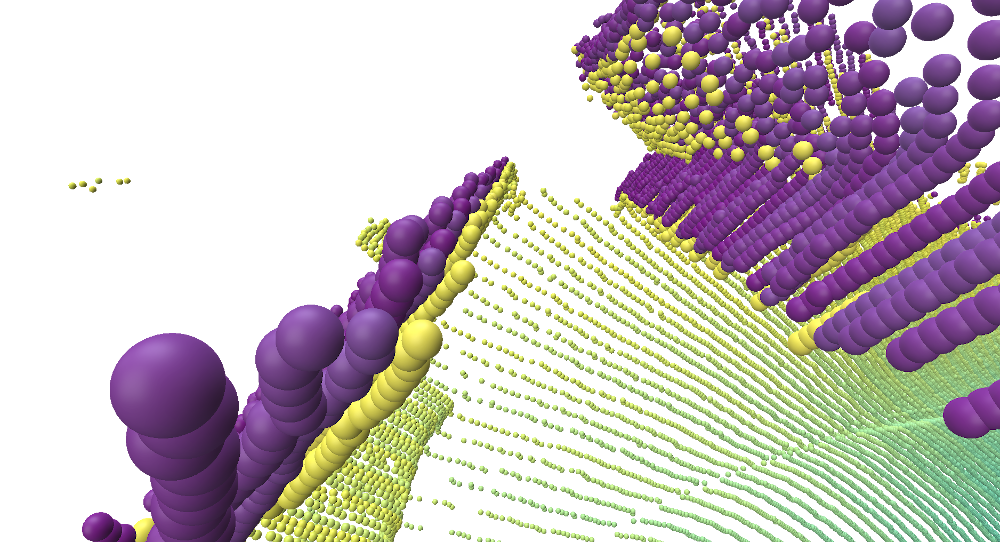}};
        \draw[color=red] (-0.95cm,0) ellipse(0.25cm and 0.8cm);
        \node[below=\legendspace of overall, text width=\imgwidth\columnwidth, align=center] {\legendsize(a) Motion-corrected lidar scan (color is time)};
        \node[below=\legendspace of ours, text width=\imgwidth\columnwidth, align=center] {\legendsize(b) \textit{Ours} motion correction without double wall};
        
        \node[below=\legendspace of overlap, text width=\imgwidth\columnwidth, align=center] {\legendsize(c) Scan self-overlap used for quantitative results};
        \node[below=\legendspace of dlio, text width=\imgwidth\columnwidth, align=center] {\legendsize(d) \textit{DLIO} motion correction with double wall};
        
    \end{tikzpicture}
    \vspace{-0.5cm}
    \caption{Illustration of the undistorting abilities of the proposed method where no ``double-wall" is visible in the motion-corrected data. (a) shows the scene's scan, (c) highlights the self-overlap area of the scan, (b) and (d) are closeups of the motion correction results of the proposed method and \textit{DLIO}~\cite{chen2023dlio} within the red ellipse in (a).}
    \label{fig:double_wall_example}
\end{figure}
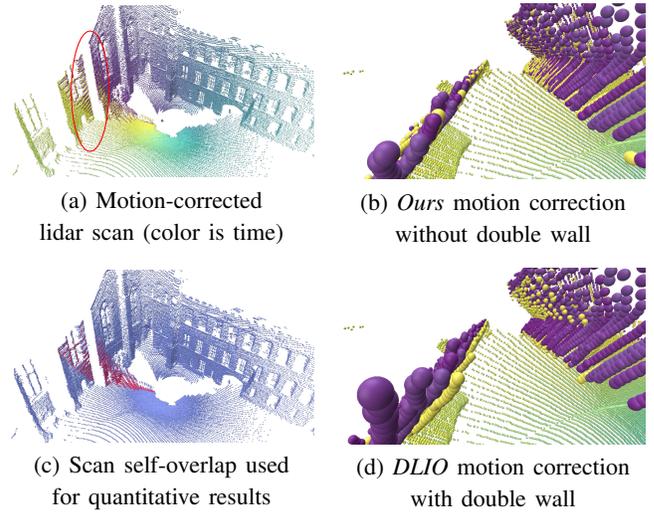

\begin{figure}
    \centering
    \includegraphics[width=1\columnwidth]{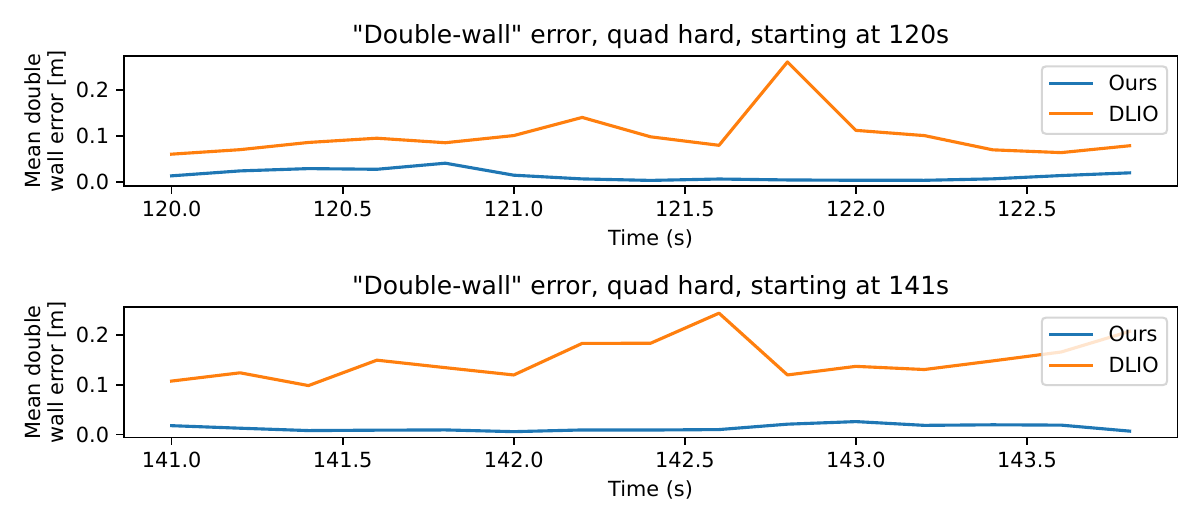}
    \vspace{-0.5cm}
    \caption{Comparison of the motion distortion correction accuracy between the proposed method and state-of-the-art lidar-inertial odometry frameworks.}
    \label{fig:double_wall_graph}
\end{figure}

Fig.~\ref{fig:double_wall_example} shows the ``double-wall" problem that occurs when the undistortion is not optimal.
In Fig.~\ref{fig:double_wall_graph}, we report the ``gap" within the double walls by computing point-to-plane distances in areas where lidar data self-overlaps between the beginning and the end of the sweep (cf. Fig.~\ref{fig:double_wall_example}(c)).
As expected, our method accurately corrects the motion distortion while \textit{DLIO} cannot close the double-wall gap as its undistortion mechanism strongly relies on the knowledge of the initial conditions.
These results demonstrate how truly-coupled methods can estimate ego-motion with higher accuracy compared to frameworks based on ``one-off" undistortion of the lidar data.

\subsubsection{Discussion}

The level of accuracy of \textit{DLIO}'s undistortion has been proven to be sufficient for many lidar-inertial odometry and mapping applications where generally the system is switched on in a relatively static position.
However, we showed that its ``on-off" undistortion mechanism reaches its limits in more challenging scenarios where the issue of ``double walls" arises.
In the context of dynamic object detection, this can lead to misdetections: a gap of $0.1\units{m}$ between surfaces observed approximately $100\units{ms}$ apart corresponds to an apparent velocity of $1\units{m/s}$ (slow walking pace).
We demonstrated that our approach allows for very accurate motion-distortion correction, thus enabling downstream applications such as dynamic object detection (cf., Section~\ref{sec:exp_dynamic}).

All our experiments run on a consumer-grade laptop with an \textit{Intel i7-1370P} CPU and $32\units{GB}$ of RAM.
All methods display a very low CPU usage with around $3\%$ for \textit{Fast-LIO2}, $6\%$ for \textit{DLIO}, and $7\%$ for ours.
It is interesting to note that the sliding-window nature of our implementation is motivated by the task of dynamic object detection that requires significant spatial overlap in the undistorted lidar data for more robust detections but still displays high level of efficiency.
For pure odometry, one can imagine even more efficient computations by removing redundant estimations.


\subsection{Dynamic object detection}
\label{sec:exp_dynamic}

In this subsection, we evaluate the integration of both our motion distortion correction and dynamic object detection through a comparison with \textit{Dynablox}~\cite{schmid2023dynablox}, a state-of-the-art non-learning dynamic object detection framework.
This method relies on maintaining a voxel-based occupancy map of the environment, detecting the dynamic points, and clustering them into objects.

To the best of our knowledge, there is no public dataset that provides raw, synchronised lidar and inertial data as well as point-wise labels for point dynamicity.
With this in mind, we chose to add the dynamicity information to the \textit{Newer College dataset} (\textit{quad\_easy}, \textit{medium}, and \textit{hard}) and we collected a novel dataset referred to as the \textit{Techlab dataset}.
The latter contained two labelled sequences that have been collected in our Techlab facility at the University of Technology Sydney with a 16-beam OS1 Ouster lidar and its embedded \ac{imu}.
A Vicon motion capture system has been used to collect the pose of the lidar relative to an earth-fixed frame.
For both datasets, the generation of the ground-truth dynamicity of each point follows a similar two-step process: the first one is mapping the environment without dynamic objects, the second one is the registration of each of the collected lidar points and query the distance between each point and the static map.
A point is classified as dynamic if the queried distance is above a certain threshold.\footnote{Some frames are not labelled due to lack of groundtruth pose. Other points are ignored, as not all the areas observed by the lidar are present in the map.}
For the \textit{Newer college dataset}, we have manually cleaned the provided map to remove dynamic objects and used the provided groundtruth trajectory for point registration.
For the \textit{Techlab dataset} the map has been create with the help of the Vicon system and the lidar data collected in an empty lab.
In the first \textit{Techlab} sequence, three people and the sensor carrier are constantly walking around the room, while in the second sequence, people momentarily stop before starting to move again.
Note that with the proposed ground-truth generation process, a person that stops will be considered as dynamic despite have a velocity equal to zero.

\begin{table}
    \centering
    \caption{Evaluation of the Intersection over Union (IoU), Recall, Accuracy, Precision, and F1 score between Dynablox~\cite{schmid2023dynablox} and our proposed method on the \textit{Techlab} and the \textit{Newer College} datasets.}
    \def\textsize{\footnotesize}
    \def\smallcol{4.1em}
    \newcolumntype{C}[1]{>{\centering\arraybackslash}p{#1}}
    \setlength{\tabcolsep}{0.2em}
    \begin{tabular}{ l l C{\smallcol} C{\smallcol} C{\smallcol} C{\smallcol} C{\smallcol}}
        \toprule
        \multicolumn{2}{l}{\scriptsize\textit{Techlab}} & \textsize IoU & \textsize Recall & \textsize Accuracy & \textsize Precision & \textsize F-1 \\
        \midrule
        \multirow{2}*{\textit{Sequence 1}}
         & \textit{Ours} & 0.61 & 0.72 & 0.99 & 0.81 & 0.76 \\ 
         & \cite{schmid2023dynablox} & 0.74 & 0.74 & 0.99 & 0.99 & 0.85 \\ 
        \midrule
        \multirow{2}*{\textit{Sequence 2}} & \textit{Ours} & 0.45 & 0.48 & 0.98 & 0.85 & 0.61 \\ 
         & \cite{schmid2023dynablox} & 0.67 & 0.67 & 0.99 & 0.99 & 0.80 \\ 
        \midrule
        \\
        \midrule
        \multicolumn{2}{l}{\scriptsize\textit{Newer College}} & \textsize IoU & \textsize Recall & \textsize Accuracy & \textsize Precision & \textsize F1 \\
        \midrule
        \multirow{2}*{\textit{quad\_easy}}
         & \textit{Ours} & 0.80 & 0.94 & 0.97 & 0.85 & 0.89 \\ 
         & \cite{schmid2023dynablox} & 0.93 & 0.94 & 0.99 & 0.99 & 0.96 \\ 
        \midrule
        \multirow{2}*{\textit{quad\_medium}} & \textit{Ours} & 0.79 & 0.89 & 0.95 & 0.87 & 0.88 \\ 
         & \cite{schmid2023dynablox} & 0.94 & 0.95 & 0.99 & 0.99 & 0.97 \\ 
         \midrule
        \multirow{2}*{\textit{quad\_hard}} & \textit{Ours} & 0.57 & 0.76 & 0.94 & 0.70 & 0.73 \\ 
         & \cite{schmid2023dynablox} & 0.90 & 0.92 & 0.99 & 0.98 & 0.95 \\ 
        \bottomrule
        \multicolumn{7}{c}{\scriptsize For all metrics, higher is better.}
    \end{tabular}
    \label{tab:quantitative_results}
\end{table}

The evaluation shown in Table~\ref{tab:quantitative_results} compares the performance of the proposed method and \textit{Dynablox} to which we provided the ground truth localisation as \textit{Dynablox} does not address the problem of scan registration/motion distortion correction.
The different metrics are computed for points less than $20\units{m}$ away from the lidar.
Across the different metrics, \textit{Dynablox} shows better results than our proposed pipeline.
However, \textit{Dynablox} implicitly leverages information from all the previous lidar scans while ours only consider the last set of undistorted lidar data (only three scans). 
Additionally, it is important to note that our method and \textit{Dynablox} have a slightly different definition for dynamic objects.
Ours looks at the instant velocity of the points while \textit{Dynablox} is detecting points that are in voxels that have previously been classified as free.
Thus, \textit{Dynablox} still considers a person that stopped moving as dynamic in the same way as the ground-truth labels are generated.
This explains why the difference between our method and \textit{Dynablox} is smaller on \textit{Techlab 1} and larger on the other data sequences.

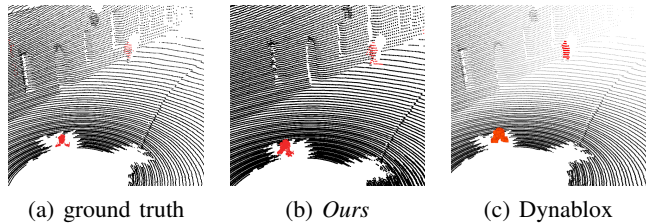
\begin{figure}
    \centering
    \def\legendsize{\small}
    \def\imgwidth{0.30}
    \def\vertspace{1cm}
    \def\horispace{0.1cm}
    \def\legendspace{-0.05cm}
    \begin{tikzpicture}
        \node (overall) {\includegraphics[clip, trim=4cm 0cm 6cm 0cm, width=\imgwidth\columnwidth]{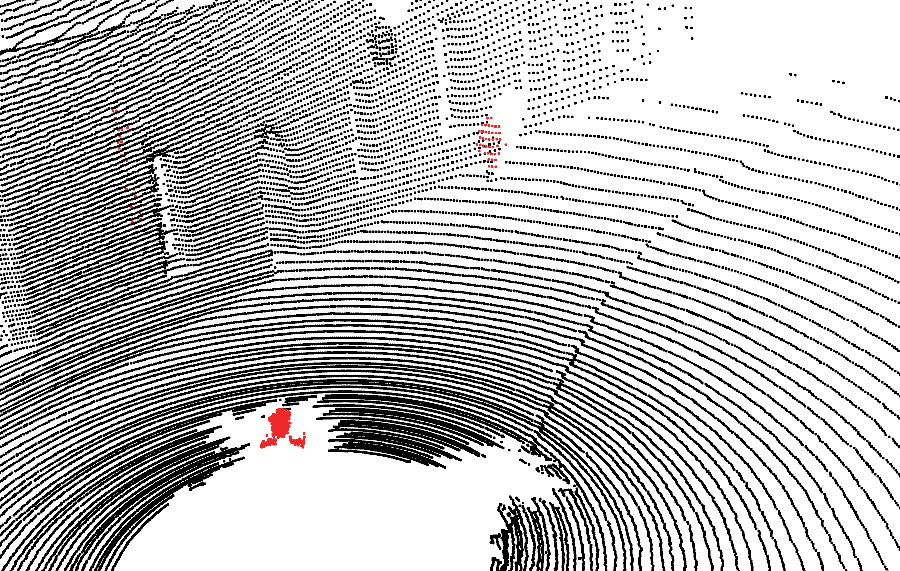}};
        \node[right=\horispace of overall] (ours) {\includegraphics[clip, trim=4cm 0cm 6cm 0cm, width=\imgwidth\columnwidth]{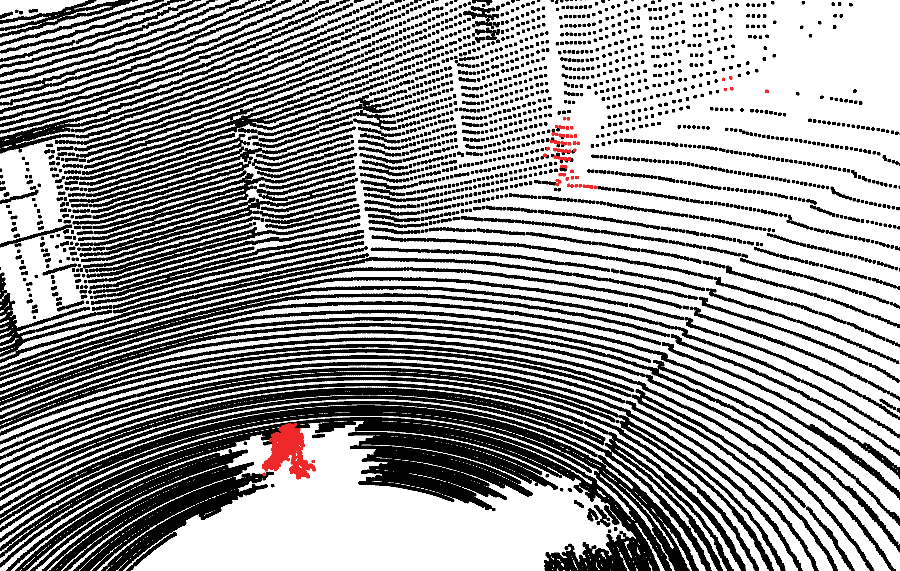}};
        \node[right = \horispace of ours] (dynablox) {\includegraphics[clip, trim=4cm 0cm 6cm 0cm, width=\imgwidth\columnwidth]{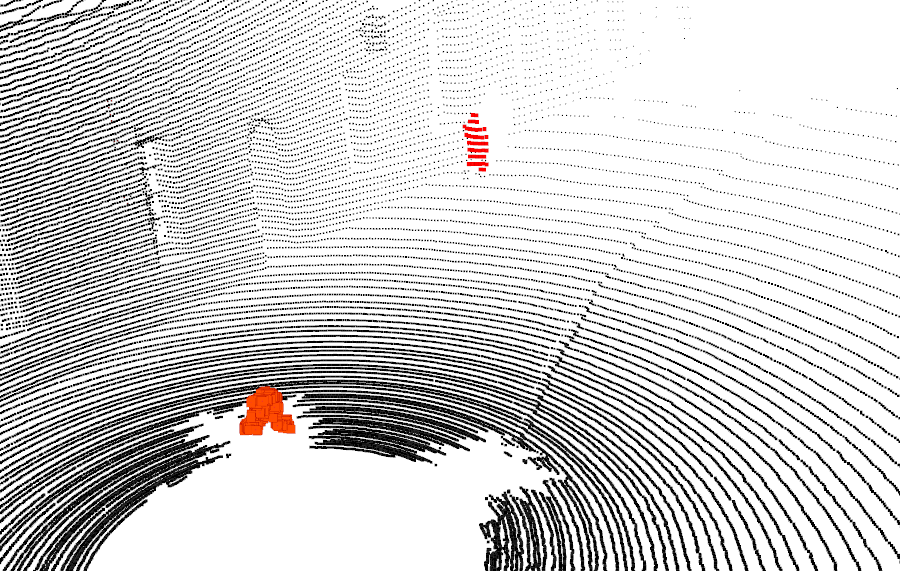}};
        
        \node[below=\legendspace of overall, text width=\imgwidth\columnwidth, align=center] {\legendsize(a) ground truth};
        \node[below=\legendspace of ours, text width=\imgwidth\columnwidth, align=center] {\legendsize(b) \textit{Ours}};
        
        \node[below=\legendspace of dynablox, text width=\imgwidth\columnwidth, align=center] {\legendsize(c) Dynablox};
        
    \end{tikzpicture}
    \caption{Illustration of the dynamic object detection. Both methods have similar performances.}
    \label{fig:dynamic}
\end{figure}

According to the different metrics and the qualitative example in Fig.~\ref{fig:dynamic}, one can see that our method performs well at detecting dynamic points but tends to classify some static points as dynamic (e.g., ground points near the pedestrian's feet).
The degradation in the performance of our method in terms of IoU, recall, and precision for \textit{quad hard} is partly explained by the imperfect ground truth as aforementioned and the fact that the dataset does not contain many dynamic points.
Nonetheless, the high accuracy allows for robust removal of dynamic elements from the lidar data.
Additionally, as the proposed method only relies on the current set of undistorted lidar data, our approach is not affected by any trajectory drift while mapping-based frameworks like \textit{Dynablox} are.
Thus, future work includes the use of our dynamic-free motion-corrected lidar data in a robust odometry and mapping framework.

\section{Conclusion}

In this paper, we presented a method for lidar-inertial motion distortion correction with application to dynamic object detection.
The first part consists of locally estimating the ego-motion of the sensors.
It is performed by coupling each lidar point with inertial information via the use of continuous \ac{imu} preintegration.
With such a formulation, the trajectory of the sensor suite is fully characterised using only 11 variables (\ac{imu} biases, initial velocity, and orientation of gravity vector).
The estimated state is the result of a non-linear least-square optimisation that aims at minimising point-to-plane and point-to-line distances.

The availability of motion-corrected lidar data allows us to perform dynamic object detection based on the spatiotemporal normals at each point.
Concretely, the lidar points are stored all together with their individual timestamps.
Similarly to standard geometric normal computation, the method queries the spatial neighbour of each point, but instead of computing the spatial covariance ($3\times3$) of each neighbourhood, our method computes the spatiotemporal covariance ($4\times4$).
By performing the eigendecomposition to find the normal vector (eigenvector with the smallest eigenvalue), we use its temporal component as a proxy for the points' velocity in space.
Our real-time implementation has been tested in real data and made open-source.

Future work includes the integration of the proposed method in a global odometry/SLAM framework to ensure geometric consistency at large timescales.
We will also explore more coupled interaction between state estimation and dynamic object detection to provide a principled approach for outlier rejection in ego-motion estimation.

\bibliographystyle{IEEEtran}
\bibliography{bibliography,sample}

\end{document}

%% file: macros.tex
\usepackage[dvipsnames]{xcolor}
\usepackage{moreverb,url}
\usepackage{amsmath,amsfonts,amssymb}
\usepackage{tikz,tkz-euclide}
\usepackage{balance}
\usetikzlibrary{decorations,calligraphy}
\usetikzlibrary{calc}
\tikzset{>=latex}

\usepackage[colorlinks,bookmarksopen,bookmarksnumbered,citecolor=red,urlcolor=red]{hyperref}
\usepackage[nolist,nohyperlinks]{acronym}

\newcommand{\units}[1]{\,\mathrm{#1}}

\renewcommand\time[1]{{t_{#1}}}
\newcommand\frametime[1]{{\tau_{#1}}}

\newcommand\refframe[2]{{\mathfrak{F}_{#1}^{#2}}}
\newcommand{\T}[2]{{\mathbf{T}_{#1}^{#2}}}
\def\Tc{{\mathbf{T}_I^L}}

\def\state{{\mathcal{S}}}

\newcommand\rot[2]{{\mathbf{R}_{#1}^{#2}}}

\newcommand\pos[2]{{\mathbf{p}_{#1}^{#2}}}

\newcommand\vel[2]{{\mathbf{v}_{#1}^{#2}}}

\def\gravity{{\mathbf{g}}}
\def\assoset{{\mathcal{A}}}
\newcommand{\asso}[1]{{\mathfrak{a}_{#1}}}

\def\acc{{f}}

\def\gyr{{\omega}}

\newcommand\biasacc[1]{{\mathbf{b}_{\acc}^{#1}}}
\newcommand\biasgyr[1]{{\mathbf{b}_{\gyr}^{#1}}}

\newcommand\dpos[2]{{\Delta\mathbf{p}_{#1}^{#2}}}

\newcommand\drot[2]{{\Delta\mathbf{R}_{#1}^{#2}}}

\def\coordinate{{\mathbf{e}}}

\newcommand\point[2]{{\mathbf{x}_{#1}^{#2}}}
\newcommand{\pointcloud}[2]{{\mathcal{P}_{#1}^{#2}}}

\newcommand\feature[2]{{\mathbf{f}_{#1}^{#2}}}

\def\nn{{\mathcal{N}}}
\def\mean{{\mathbf{m}}}
\def\cov{{\text{cov}}}

\def\dscore{{s}}

\begin{acronym}[AAAAAAAAA]
    \acro{1d}[1D]{One-Dimensional}
    \acro{2d}[2D]{Two-Dimensional}
    \acro{3d}[3D]{Three-Dimensional}
    \acro{cas}[CAS]{Centre for Autonomous Systems}
    \acro{cnn}[CNN]{Convolutional Neural Network}
    \acro{cpu}[CPU]{Central Processing Unit}
    \acro{doals}[DOALS]{Urban Dynamic Objects LiDAR}
    \acro{dof}[DoF]{Degree-of-Freedom}
    \acro{dvs}[DVS]{Dynamic Vision Sensor}
    \acrodefplural{dvs}[DVS's]{Dynamic Vision Sensors}
    \acro{ekf}[EKF]{Extended Kalman filter}
    \acro{fifo}[FIFO]{First In, First Out}
    \acro{fov}[FoV]{Field-of-View}
    \acro{gnss}[GNSS]{Global Navigation Satellite System}
    \acrodefplural{gnss}[GNSS's]{Global Navigation Satellite Systems}
    \acro{gp}[GP]{Gaussian Process}
    \acrodefplural{gp}[GPs]{Gaussian Processes}
    \acro{gpm}[GPM]{Gaussian Preintegrated Measurement}
    \acro{ugpm}[UGPM]{Unified Gaussian Preintegrated Measurement}
    \acro{gps}[GPS]{Global Position System}
    \acrodefplural{gps}[GPS's]{Global Position Systems}
    \acro{gpu}[GPU]{Graphic Processing Unit}
    \acro{hdr}[HDR]{High Dynamic Range}
    \acro{hri}[HRI]{Human-Robot Interactions}
    \acro{icp}[ICP]{Iterative Closest Point}
    \acro{idol}[IDOL]{IMU-DVS Odometry using Lines}
    \acro{iou}[IoU]{Intersection over Union}
    \acro{imu}[IMU]{Inertial Measurement Unit}
    \acro{in2laama}[IN2LAAMA]{INertial Lidar Localisation Autocalibration And MApping}
    \acro{kf}[KF]{Kalman Filter}
    \acro{kl}[KL]{Kullback–Leibler}
    \acro{lidar}[LiDAR]{Light Detection And Ranging Sensor}
    \acro{lpm}[LPM]{Linear Preintegrated Measurement}
    \acro{map}[MAP]{Maximum A Posteriori}
    \acro{mle}[MLE]{Maximum Likelihood Estimation}
    \acro{ndt}[NDT]{Normal Distribution Transform}
    \acro{pca}[PCA]{Principal Component Analysis}
    \acro{pm}[PM]{Preintegrated Measurement}
    \acro{rcnn}[R-CNN]{Region-based Convolutional Neural Network}
    \acro{rrbt}[RRBT]{Rapidly Exploring Random Belief Trees}
    \acro{rgb}[RGB]{color}
    \acro{rgbd}[RGBD]{color-depth}
    \acro{rms}[RMS]{Root Mean Squared}
    \acro{rmse}[RMSE]{Root Mean Squared Error}
    \acro{ros}[ROS]{Robot Operating System}
    \acro{sde}[SDE]{Stochastic Differential Equation}
    \acro{SE3}[SE(3)]{Special Euclidean group in three dimensions}
    \acro{slam}[SLAM]{Simultaneous Localisation And Mapping}
    \acro{so3}[$\mathfrak{so}$(3)]{Lie algebra of special orthonormal group in three dimensions}
    \acro{SO3}[SO(3)]{Special Orthonormal rotation group in three dimensions}
    \acro{upm}[UPM]{Upsampled-Preintegrated-Measurement}
    \acro{uts}[UTS]{University of Technology, Sydney}
    \acro{vi}[VI]{Visual-Inertial}
    \acro{vio}[VIO]{Visual-Inertial Odometry}
    \acro{vo}[VO]{Visual Odometry}

\end{acronym}

%% file: figures/overview_undistortion.tex
\def\hdist{6em}
\def\vdist{2.5em}
\def\blockheight{4em}
\def\blockwidth{8.5em}
\def\innerpad{0.1em}
\def\textsize{\scriptsize}
\begin{tikzpicture}[auto]
    \tikzstyle{input} = [draw, fill=white, rectangle, minimum height = 2.4em, text width = 5.5em,  minimum width = 5.5em, align = center, node distance = 5em, draw=red, execute at begin node=\setlength{\baselineskip}{8pt}]
    \tikzstyle{block} = [draw, fill=white, rectangle, minimum height = \blockheight, text width = \blockwidth,  minimum width = \blockwidth, align = center, inner sep=\innerpad, outer sep=0, node distance = 11em, execute at begin node=\setlength{\baselineskip}{8pt}] 
    \tikzstyle{largeblock} = [draw, fill=white, rectangle, minimum height = \blockheight, text width =2*\innerpad + 2*\blockwidth+\hdist,  minimum width = 2*\innerpad + 2*\blockwidth+\hdist, align = center, node distance = 11em, execute at begin node=\setlength{\baselineskip}{8pt}] 
    \tikzstyle{output} = [draw=none, fill=white, text=NavyBlue, rectangle, minimum height = 1em, text width = 2*\innerpad + 2*\blockwidth+\hdist,  minimum width = 2*\innerpad + 2*\blockwidth+\hdist, align = center, node distance = 11em, execute at begin node=\setlength{\baselineskip}{8pt}] 

    \node [input] (imu) {\textsize \textbf{6-DoF IMU}};
    \node [block, below = \vdist of imu] (preint) {\textsize \textbf{LPMs preintegration}\\Piece-wise linear continuous latent state};
    \node [block, right= \hdist of preint] (feature) {\textsize \textbf{Feature point extraction}\\Point-to-neighbour-line roughness score};
    \node [block, below= \vdist of feature] (asso) {\textsize \textbf{Data association}\\Closest neighbour search};
    \node [input, above=\vdist of feature] (lidar) {\textsize \textbf{3D lidar}};
    
    \coordinate (mid) at ($(preint.south)!0.5!(feature.south)$);
    \node [largeblock, below=2*\vdist+\blockheight of mid] (opti) {\textsize \textbf{Non-linear least-square optimisation}\\Point-to-line and point-to-plane residuals. Continuous trajectory parameterised by IMU biases, initial velocity and gravity alignment};
    
    \draw[->] (imu) -- node[text width = 5em, execute at begin node=\setlength{\baselineskip}{7pt}]{\textsize Accelerometer and gyroscope measurements} (preint);
    \draw[->] (lidar) -- node[left, align=right, text width = 3em, execute at begin node=\setlength{\baselineskip}{7pt}]{\textsize Point stream} (feature);
    \draw[->] (feature) -- node[align=center, text width = \hdist - 0.5em, execute at begin node=\setlength{\baselineskip}{7pt}]{\textsize Feature timestamps} (preint);
    \draw[->] (feature) -- node[left, align=right, text width = 4em, execute at begin node=\setlength{\baselineskip}{7pt}]{\textsize Edge and planar feature points} (asso);

    \draw[->] (preint) |- node[above right, text width = 3em, execute at begin node=\setlength{\baselineskip}{7pt}]{\textsize Preintegrated measurements} (asso);

    \coordinate[below=\vdist of asso] (optianchor);
    \draw[->] (asso) -- node[left, align=right, text width = 8em, execute at begin node=\setlength{\baselineskip}{7pt}]{\textsize Feature associations and corresponding preintegrated measurements} (optianchor);

    \node[output, below=0.5*\vdist of opti] (out) {\textsize \textbf{Motion-corrected lidar data}};
    \draw[->] (opti) -- (out);
    

\end{tikzpicture}

%% file: figures/spatiotemporal_view.tex
    \def\scale{0.6}
    \def\vertsize{4.0cm}
    
    \begin{tikzpicture}
        \node (titleaa){\scriptsize 1D spatial view (20 scans)};
        \node[below=0.0cm of titleaa] (onedspace){\includegraphics[clip, width=\scale\columnwidth, trim=1cm 9cm 1cm 2.3cm]{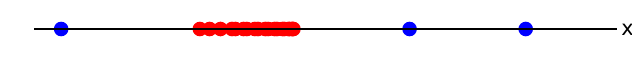}};
        \node[below=0.0cm of onedspace] (onedlegend){\includegraphics[clip, width=\scale\columnwidth, trim=1.5cm 6.8cm 1.2cm 5cm]{figures/intuition_1d.pdf}};
        \node[below=0.0cm of onedlegend] (titleab){\scriptsize 2D spatiotemporal view (20 scans)};
        \node[below=0.0cm of titleab] (onedtime){\includegraphics[clip, width=\scale\columnwidth, trim=0cm 0.5cm 0cm 7cm]{figures/intuition_1d.pdf}};
        \node[below=0.1cm of onedtime] (subfiga){(a) 1D geometric data};

        \node[right=2.5cm of titleaa] (titleba){\scriptsize 2D spatial view (20 scans)};
        \node[below=0.0cm of titleba] (twodspace){\includegraphics[clip, width=0.6\columnwidth, trim=1cm 9.65cm 1cm 0.75cm]{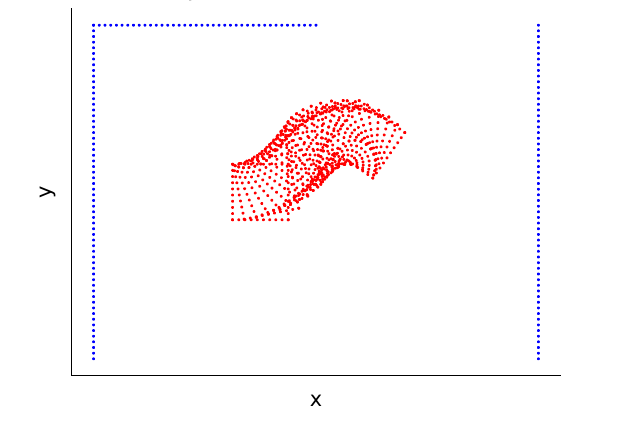}};

        \node[right=1.5cm of titleba] (titlebb){\scriptsize 3D spatiotemporal view (20 scans)};
        \node[below=0.0cm of titlebb] (twodtime){\includegraphics[clip, height=\vertsize, trim=3.5cm 0.3cm 2.8cm 11.5cm]{figures/intuition_2d.pdf}};
        \node[right=5cm of subfiga] (subfigb){(b) 2D geometric data};
        \node[below=-0.1cm of twodspace, xshift=2.8cm] (twodlegend){\includegraphics[clip, width=0.9\columnwidth, trim=1.3cm 8.5cm 1cm 8.5cm]{figures/intuition_2d.pdf}};
    \end{tikzpicture}

%% file: root.bbl
\begin{thebibliography}{10}
\providecommand{\url}[1]{#1}
\csname url@samestyle\endcsname
\providecommand{\newblock}{\relax}
\providecommand{\bibinfo}[2]{#2}
\providecommand{\BIBentrySTDinterwordspacing}{\spaceskip=0pt\relax}
\providecommand{\BIBentryALTinterwordstretchfactor}{4}
\providecommand{\BIBentryALTinterwordspacing}{\spaceskip=\fontdimen2\font plus
\BIBentryALTinterwordstretchfactor\fontdimen3\font minus \fontdimen4\font\relax}
\providecommand{\BIBforeignlanguage}[2]{{%
\expandafter\ifx\csname l@#1\endcsname\relax
\typeout{** WARNING: IEEEtran.bst: No hyphenation pattern has been}%
\typeout{** loaded for the language `#1'. Using the pattern for}%
\typeout{** the default language instead.}%
\else
\language=\csname l@#1\endcsname
\fi
#2}}
\providecommand{\BIBdecl}{\relax}
\BIBdecl

\bibitem{sunderhauf2018limits}
N.~Sünderhauf, O.~Brock, W.~Scheirer, R.~Hadsell, D.~Fox, J.~Leitner, B.~Upcroft, P.~Abbeel, W.~Burgard, M.~Milford, and P.~Corke, ``The limits and potentials of deep learning for robotics,'' \emph{The International Journal of Robotics Research}, vol.~37, no. 4-5, pp. 405--420, 2018.

\bibitem{willers2020safety}
O.~Willers, S.~Sudholt, S.~Raafatnia, and S.~Abrecht, ``Safety concerns and mitigation approaches regarding the use of deep learning in safety-critical perception tasks,'' in \emph{Computer Safety, Reliability, and Security. SAFECOMP 2020 Workshops}, A.~Casimiro, F.~Ortmeier, E.~Schoitsch, F.~Bitsch, and P.~Ferreira, Eds.\hskip 1em plus 0.5em minus 0.4em\relax Cham: Springer International Publishing, 2020, pp. 336--350.

\bibitem{lee2024lidar}
D.~Lee, M.~Jung, W.~Yang, and A.~Kim, ``Lidar odometry survey: recent advancements and remaining challenges,'' \emph{Intelligent Service Robotics}, pp. 1--24, 2024.

\bibitem{Ye2019}
H.~Ye, Y.~Chen, and M.~Liu, ``{Tightly coupled 3D Lidar inertial odometry and mapping},'' \emph{Proceedings - IEEE International Conference on Robotics and Automation}, vol. 2019-May, pp. 3144--3150, 2019.

\bibitem{shan2020liosam}
T.~Shan, B.~Englot, D.~Meyers, W.~Wang, C.~Ratti, and D.~Rus, ``Lio-sam: Tightly-coupled lidar inertial odometry via smoothing and mapping,'' in \emph{2020 IEEE/RSJ International Conference on Intelligent Robots and Systems (IROS)}, 2020, pp. 5135--5142.

\bibitem{legentil2023latent}
C.~{Le Gentil} and T.~Vidal-Calleja, ``Continuous latent state preintegration for inertial-aided systems,'' \emph{The International Journal of Robotics Research}, vol.~42, no.~10, pp. 874--900, 2023.

\bibitem{falque2023dynamic}
R.~Falque, C.~{Le Gentil}, and F.~Sukkar, ``Dynamic object detection in range data using spatiotemporal normals,'' in \emph{Australasian Conference on Robotics and Automation}, 2023.

\bibitem{schmid2023dynablox}
L.~Schmid, O.~Andersson, A.~Sulser, P.~Pfreundschuh, and R.~Siegwart, ``Dynablox: Real-time detection of diverse dynamic objects in complex environments,'' \emph{IEEE Robotics and Automation Letters}, vol.~8, no.~10, pp. 6259--6266, 2023.

\bibitem{Bosse2012}
M.~Bosse, R.~Zlot, and P.~Flick, ``{Zebedee : Design of a spring-mounted 3-D range sensor with application to mobile mapping},'' \emph{IEEE Transactions on Robotics}, vol.~28, no. October, pp. 1--15, 2012.

\bibitem{Zhang2015}
J.~Zhang and S.~Singh, ``{Visual-lidar Odometry and Mapping: Low-drift, Robust, and Fast},'' \emph{IEEE International Conference on Robotics and Automation}, pp. 2174--2181, 2015.

\bibitem{xu2022fastlio2}
W.~Xu, Y.~Cai, D.~He, J.~Lin, and F.~Zhang, ``Fast-lio2: Fast direct lidar-inertial odometry,'' \emph{IEEE Transactions on Robotics}, vol.~38, no.~4, pp. 2053--2073, 2022.

\bibitem{chen2023dlio}
K.~Chen, R.~Nemiroff, and B.~T. Lopez, ``Direct lidar-inertial odometry: Lightweight lio with continuous-time motion correction,'' in \emph{2023 IEEE International Conference on Robotics and Automation (ICRA)}, 2023, pp. 3983--3989.

\bibitem{Tang2019}
T.~Y. Tang, D.~J. Yoon, and T.~D. Barfoot, ``A white-noise-on-jerk motion prior for continuous-time trajectory estimation on <italic>se(3)</italic>,'' \emph{IEEE Robotics and Automation Letters}, vol.~4, no.~2, pp. 594--601, 2019.

\bibitem{cioffi2022continuous}
G.~Cioffi, T.~Cieslewski, and D.~Scaramuzza, ``Continuous-time vs. discrete-time vision-based slam: A comparative study,'' \emph{IEEE Robotics and Automation Letters}, vol.~7, no.~2, pp. 2399--2406, 2022.

\bibitem{Lupton2012}
T.~Lupton and S.~Sukkarieh, ``{Visual-inertial-aided navigation for high-dynamic motion in built environments without initial conditions},'' \emph{IEEE Transactions on Robotics}, vol.~28, no.~1, pp. 61--76, 2012.

\bibitem{LeGentil2018}
C.~{Le Gentil}, T.~Vidal-Calleja, and S.~Huang, ``{3D Lidar-IMU Calibration based on Upsampled Preintegrated Measurements for Motion Distortion Correction},'' \emph{IEEE International Conference on Robotics and Automation}, 2018.

\bibitem{LeGentil2021}
------, ``{IN2LAAMA: INertial Lidar Localisation Autocalibration And MApping},'' \emph{IEEE Transactions on Robotics}, 2021.

\bibitem{LeGentil2021b}
C.~{Le Gentil} and T.~Vidal-Calleja, ``{Continuous Integration over SO(3) for IMU Preintegration},'' \emph{Robotics: Science and Systems XVII}, 2021.

\bibitem{chen2019suma++}
X.~Chen, A.~Milioto, E.~Palazzolo, P.~Giguere, J.~Behley, and C.~Stachniss, ``Suma++: Efficient lidar-based semantic slam,'' in \emph{2019 IEEE/RSJ International Conference on Intelligent Robots and Systems (IROS)}.\hskip 1em plus 0.5em minus 0.4em\relax IEEE, 2019, pp. 4530--4537.

\bibitem{milioto2019rangenet++}
A.~Milioto, I.~Vizzo, J.~Behley, and C.~Stachniss, ``Rangenet++: Fast and accurate lidar semantic segmentation,'' in \emph{2019 IEEE/RSJ international conference on intelligent robots and systems (IROS)}.\hskip 1em plus 0.5em minus 0.4em\relax IEEE, 2019, pp. 4213--4220.

\bibitem{geiger2012we}
A.~Geiger, P.~Lenz, and R.~Urtasun, ``Are we ready for autonomous driving? the kitti vision benchmark suite,'' in \emph{2012 IEEE conference on computer vision and pattern recognition}.\hskip 1em plus 0.5em minus 0.4em\relax IEEE, 2012, pp. 3354--3361.

\bibitem{chen2021moving}
X.~Chen, S.~Li, B.~Mersch, L.~Wiesmann, J.~Gall, J.~Behley, and C.~Stachniss, ``Moving object segmentation in 3d lidar data: A learning-based approach exploiting sequential data,'' \emph{IEEE Robotics and Automation Letters}, vol.~6, no.~4, pp. 6529--6536, 2021.

\bibitem{henein2020dynamic}
M.~Henein, J.~Zhang, R.~Mahony, and V.~Ila, ``Dynamic slam: The need for speed,'' in \emph{2020 IEEE International Conference on Robotics and Automation (ICRA)}.\hskip 1em plus 0.5em minus 0.4em\relax IEEE, 2020, pp. 2123--2129.

\bibitem{oleynikova2017voxblox}
H.~Oleynikova, Z.~Taylor, M.~Fehr, R.~Siegwart, and J.~Nieto, ``Voxblox: Incremental 3d euclidean signed distance fields for on-board mav planning,'' in \emph{IEEE/RSJ International Conference on Intelligent Robots and Systems (IROS)}, 2017.

\bibitem{mersch2023ral}
M.~Benedikt, G.~Tiziano, C.~Xieyuanli, V.~Ignacio, B.~Jens, and S.~Cyrill, ``{Building Volumetric Beliefs for Dynamic Environments Exploiting Map-Based Moving Object Segmentation},'' \emph{IEEE Robotics and Automation Letters (RA-L)}, vol.~8, no.~8, pp. 5180--5187, 2023.

\bibitem{Bentley1975}
J.~L. Bentley, ``{Multidimensional binary search trees used for associative searching},'' \emph{Communications of the ACM}, vol.~18, no.~9, pp. 509--517, sep 1975.

\bibitem{legentil2024reverting}
C.~Le~Gentil, O.-L. Ouabi, L.~Wu, C.~Pradalier, and T.~Vidal-Calleja, ``Accurate gaussian-process-based distance fields with applications to echolocation and mapping,'' \emph{IEEE Robotics and Automation Letters}, vol.~9, no.~2, pp. 1365--1372, 2024.

\bibitem{Agarwal}
\BIBentryALTinterwordspacing
S.~Agarwal and K.~Mierle, ``{Ceres Solver}.'' [Online]. Available: \url{http://ceres-solver.org}
\BIBentrySTDinterwordspacing

\bibitem{zhang2021multicamera}
L.~Zhang, M.~Camurri, D.~Wisth, and M.~Fallon, ``Multi-camera lidar inertial extension to the newer college dataset,'' 2021.

\end{thebibliography}
